\documentclass[letterpaper, 10 pt, journal, twoside]{IEEEtran} 

\IEEEoverridecommandlockouts                              

\PassOptionsToPackage{export}{adjustbox}
\usepackage{definitions}
\usepackage{xcolor}
\usepackage[T1]{fontenc}
\usepackage{tikz}
\usepackage{array}
\usepackage{multirow}

\makeatletter
\newcommand{\shorteq}{
	\settowidth{\@tempdima}{-}
	\resizebox{\@tempdima}{\height}{=}
}
\makeatother

\definecolor{mGreen}{rgb}{0.36, 0.69, 0}
\newcommand{\dso}{{\mathds{1}}}
\newcommand{\cps}{{\underline{\mathscr{p}}}}
\newcommand{\cpr}{{\underline{\mathscr{q}}}}
\newcommand{\cpb}{{\underline{\mathscr{b}}}}
\renewcommand{\ur}{\underline{q}}
\renewcommand{\us}{\underline{p}}
\newcommand{\uph}{{\underline{\phi}}}

\newcommand{\ude}{{\underline{\delta}}}

\newcommand{\scv}{{\mathscr{v}}}

\newcommand{\scF}{{\mathscr{F}}}

\newcommand{\twi}{{\text{w}}}
\newcommand{\ez}{{\mathscr{\uz}}}
\newcommand{\hua}{{\hat{\ua}}}
\newcommand{\huo}{{\hat{\uom}}}
\newcommand{\cpd}{{\underline{\mathscr{d}}}}
\newcommand{\cpx}{{\underline{\mathscr{x}}}}
\newcommand{\cpX}{{\mathcal{{X}}}}
\newcommand{\qf}{{\mathscr{Q}}}

\newcommand{\ttv}{{\text{v}}}
\newcommand{\Dt}{{\Delta}}
\newcommand{\ind}{{\mathbb{I}}}
\newcommand{\scR}{{\mathscr{R}}}
\newcommand{\die}{{\underline{\mathscr{e}}}}
\newcommand{\eva}[1]{\bigg\vert_{#1}}
\newcommand{\ttt}[1]{{\texttt{#1}}}
\newcommand{\uwb}{{\ttt{u}}}
\newcommand{\imu}{{\ttt{i}}}

\newcommand{\hzz}{{\hat{z}}}

\newcommand{\acc}{{\ttt{acc}}}
\newcommand{\gyro}{{\ttt{gyro}}}

\newcommand{\bias}{{\ttt{bias}}}
\newcommand{\ug}{{\underline{g}}}
\newcommand{\bo}{{\bm\Omega}}
\newcommand{\tta}{{\ttt{ta}}}
\newcommand{\ttn}{{\ttt{an}}}

\newcommand{\fimu}{{\ttt{I}}}
\newcommand{\fuwb}{{\ttt{U}}}
\newcommand{\fnav}{{\ttt{W}}}

\newcommand{\rightArrow}[1]{\parbox{#1}{\tikz{\draw[->](0,0)--(#1,0);}}}
\newcommand{\leftArrow}[1]{\parbox{#1}{\tikz{\draw[<-](0,0)--(#1,0);}}}
\newcommand{\ra}{{\rightArrow{.15cm}}}
\newcommand{\la}{{\leftArrow{.15cm}}}

\newcommand{\sn}[2]{{{#1}\times10^{#2}}}

\DeclareCaptionLabelFormat{andtable}{#1~#2  \&  \tablename~\thetable}

\def\BibTeX{{\rm B\kern-.05em{\sc i\kern-.025em b}\kern-.08em
		T\kern-.1667em\lower.7ex\hbox{E}\kern-.125emX}}
\usepackage{balance}

\newcommand{\red}[1]{{#1}}

\begin{document}

	\markboth{Published on IEEE Robotics and Automation Letters (RA-L). DOI: 10.1109/LRA.2023.3281932. \textcopyright IEEE}{}

	\author{Kailai Li$^{1}$, Ziyu Cao$^{2}$, and Uwe D. Hanebeck$^{2}$
	\thanks{Published on IEEE Robotics and Automation Letters (RA-L). \textcopyright 2023 IEEE. Personal use of this material is permitted. Permission from IEEE must be obtained for all other uses. DOI: 10.1109/LRA.2023.3281932}
	\thanks{This work was partially supported by the German Federal Ministry of Education and Research (BMBF) under grant POMAS (FKZ 01IS17042), the German Research Foundation (DFG) under grant HA 3789/25-1, and the Swedish Research Council under grant Scalable Kalman Filters.}
	\thanks{$^{1}$Kailai Li is with the Division of Automatic Control, Department of Electrical Engineering, Linköping University, Sweden (e-mail: {\tt\footnotesize kailai.li@liu.se})}
	\thanks{$^{2}$Ziyu Cao and Uwe D. Hanebeck are with the Intelligent Sensor-Actuator-Systems Laboratory, Institute for Anthropomatics and Robotics, Karlsruhe Institute of Technology, Germany (e-mail: {\tt\footnotesize ziyu.cao@kit.edu}; {\tt\footnotesize uwe.hanebeck@kit.edu})}
	}
	
	\title{Continuous-Time Ultra-Wideband-Inertial Fusion}
		
	\maketitle
	\begin{abstract}
		We introduce a novel framework of continuous-time ultra-wideband-inertial sensor fusion for online motion estimation. Quaternion-based cubic cumulative B-splines are exploited for parameterizing motion states continuously over time. Systematic derivations of analytic kinematic interpolations and spatial differentiations are further provided. Based thereon, a new sliding-window spline fitting scheme is established for asynchronous multi-sensor fusion and online calibration. We conduct a dedicated validation of the quaternion spline fitting method, and evaluate the proposed system, SFUISE (spline fusion-based ultra-wideband-inertial state estimation), in real-world scenarios using public data set and experiments. The proposed sensor fusion system is real-time capable and delivers superior performance over state-of-the-art discrete-time schemes. We release the source code and own experimental data at \texttt{https://github.com/KIT-ISAS/SFUISE}\,.
	\end{abstract}

	%--------------------------------------------------------------------------------------
	
	\section{Introduction and Related Work}\label{sec:introduction}
	\IEEEPARstart{O}{nline} estimation of dynamical motions is of fundamental importance in achieving reliable autonomy of mobile robots~\cite{RAL21_Li,leutenegger2015keyframe,ICRA20_Li,li2020mine}. Recent advancements in ultra-wideband (UWB) technology have offered promising alternative solutions to localization in GPS-denied environments. Compared with common sensing principles, e.g., cameras or LiDARs, UWB sensors are lightweight, low-cost, and more scalable in large-scale deployment, particularly, in indoor scenarios~\cite{zheng2022uwb}.	Despite the high spatial resolution of ultra-wideband impulse radio, which transmits at the nanosecond level, there are still technical challenges to achieving high-performance UWB-based tracking in practice. Non-line-of-sight (NLOS) and multipath conditions are well-known issues in UWB ranging, which are dependent on sensor placements and can be further exacerbated by complex and time-varying environments, such as those with moving obstacles~\cite{zhao2022finding,zhao2022util}. In addition, UWB ranging often exhibits non-Gaussian noise patterns, with surrounding-dependent interference and diffraction that are impossible to model parametrically~\cite{hol2009uwb,kok2015uwb,zhao2021learning}. Thus, basic UWB tracking solutions using multilateration are almost always insufficient for high-performance motion estimation.
	\begin{figure}[t]
		\centering
		\begin{tabular}{cc}
			\adjustbox{trim={0.2\width} {0.19\height} {0.15\width} {.24\height},clip}{\includegraphics[width=0.34\textwidth]{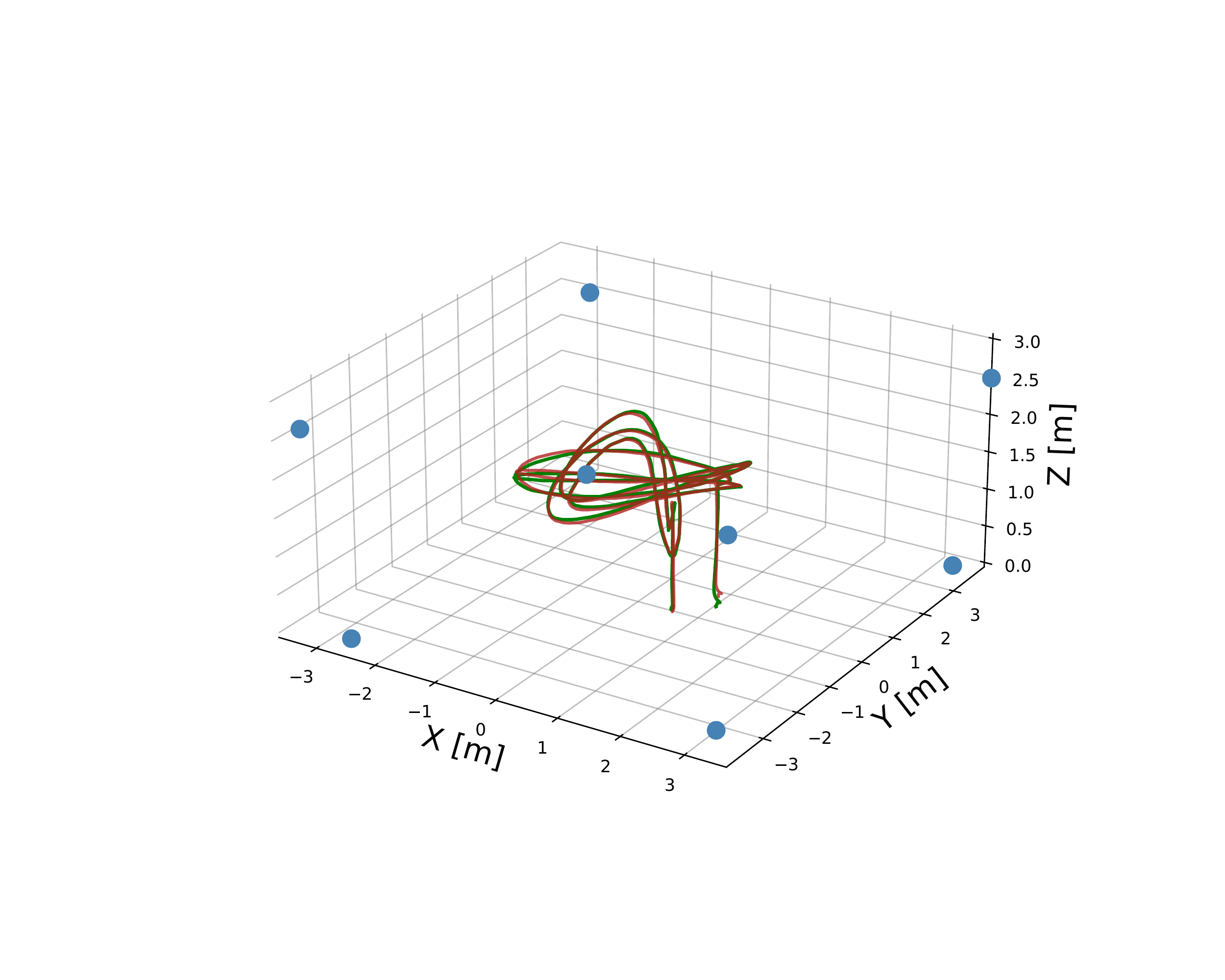}}&
			\adjustbox{trim={0.2\width} {0.19\height} {0.15\width} {.24\height},clip}{\includegraphics[width=0.34\textwidth]{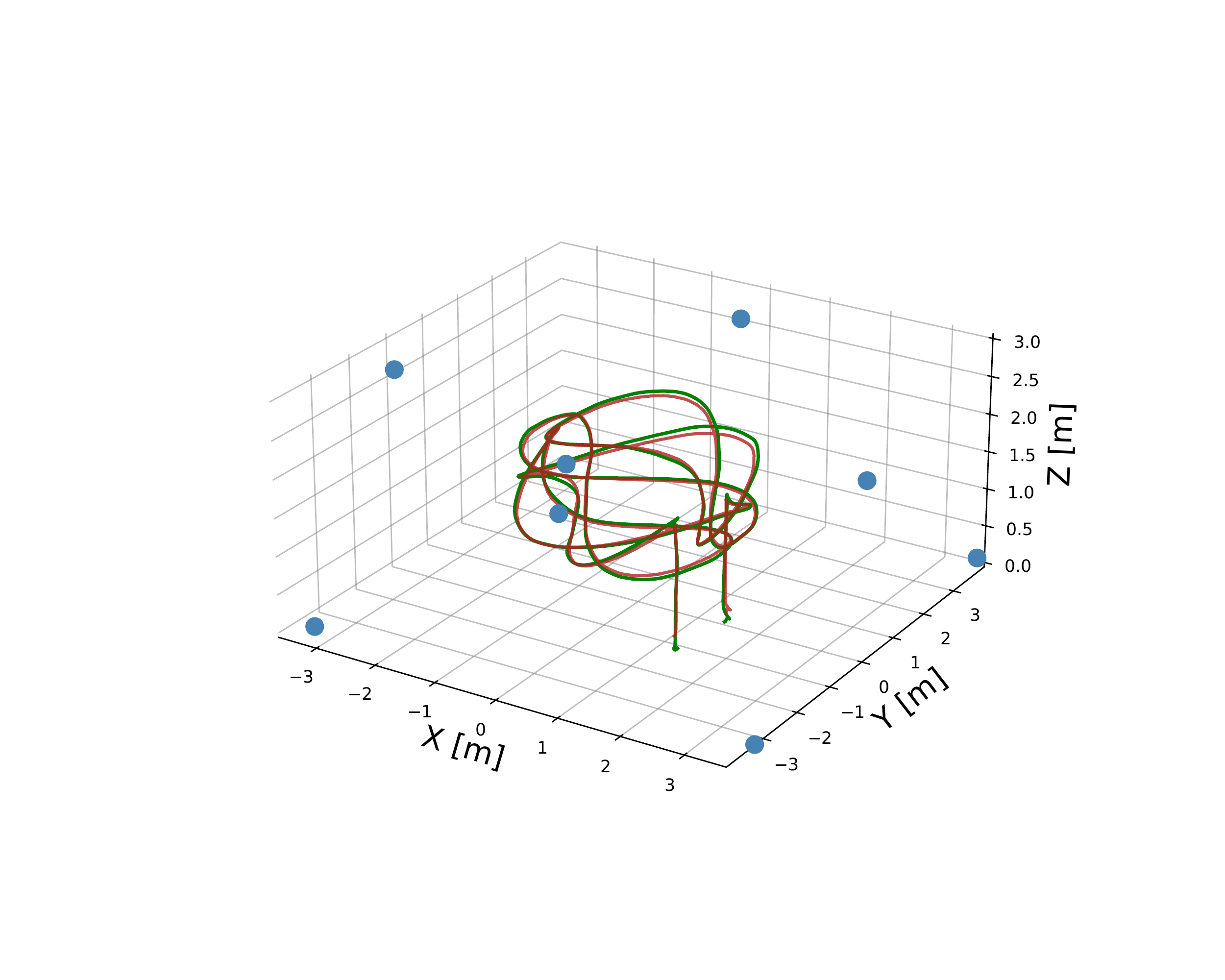}}\\
			\subcap{(A) \ttt{const1-trial6-tdoa2}} &\subcap{(B) \ttt{const2-trial3-tdoa3}}\\
			\adjustbox{trim={0.2\width} {0.19\height} {0.15\width} {.24\height},clip}{\includegraphics[width=0.34\textwidth]{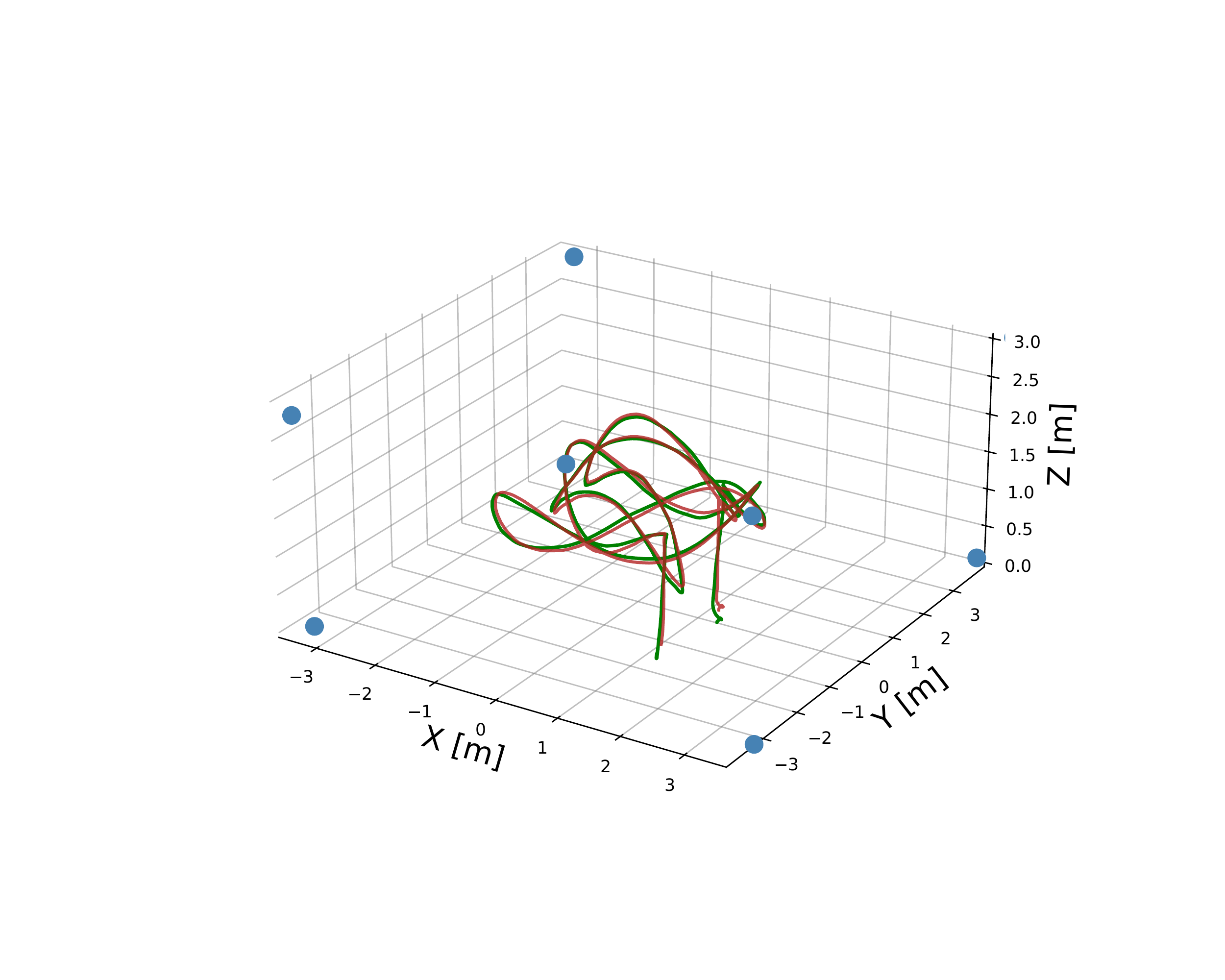}}&
			\adjustbox{trim={0.2\width} {0.19\height} {0.15\width} {.24\height},clip}{\includegraphics[width=0.34\textwidth]{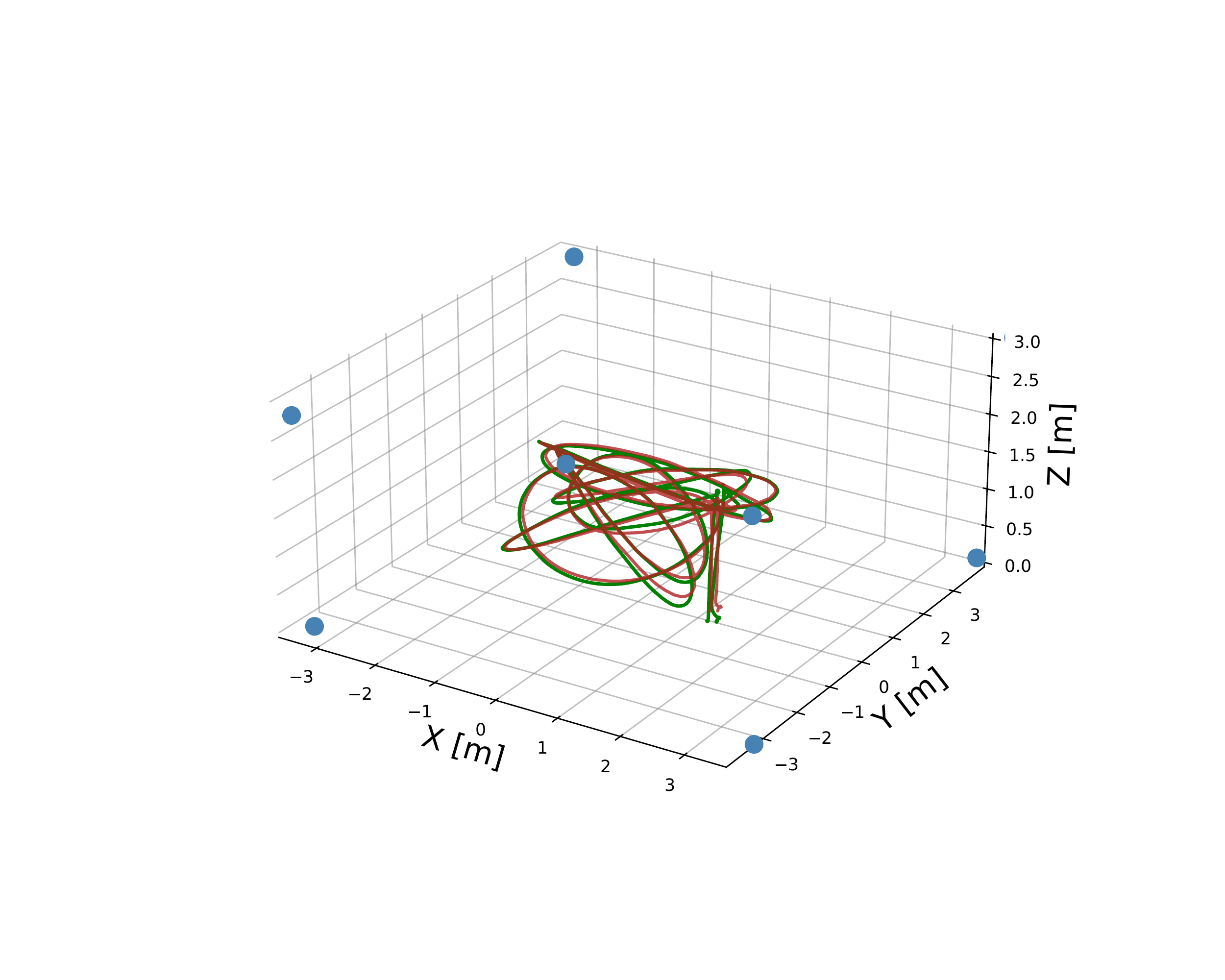}}\\
			\subcap{(C) \ttt{const3-trial4-tdoa3}} &\subcap{(D) \ttt{const3-trial5-tdoa3}}
		\end{tabular}
		\caption{Exemplary runs of SFUISE on \ttt{UTIL}. The proposed system delivers accurate trajectory estimates (green) compared to ground truth (red). Blue dots depict anchor positions.}
		\label{fig:sfuise}
	\end{figure}

    The performance of UWB tracking can be improved through sensor fusion with other modalities. \red{Inertial measurement units (IMUs)} provide instantaneous and higher-order motion information that can bridge the gap between consecutive UWB readings. They are cost- and resource-efficient, and can be easily integrated into UWB sensor networks. Conventional UWB-inertial fusion methods often rely on recursive filtering algorithms, particularly the extended Kalman filter (EKF), with inertial measurements facilitating state propagation and ultra-wideband ranging updating the predicted prior. One basic application was introduced in~\cite{liG2018uwb}, where an EKF was used to localize micro aerial vehicles with a UWB-inertial setup. Further practices involve utilizing six-DoF motion estimation through quaternion kinematics, and incorporating error-states to enhance overall system performance~\cite{hol2009uwb,sola2017quaternion,goudar2021uwb}.
    
    However, recursive filters rely on the Markov assumption, where evidence for predicting the current state is only traced back to the last state~\cite{li2020mine}. Sensor measurements of different modalities are usually fused in a decoupled manner, inducing substantial information loss in correlations across multi-sensor readings~\cite{leutenegger2015keyframe}. These issues can be substantially mitigated by fusing sensor measurements into one joint graph-based nonlinear optimization. It improves tracking accuracy, while still maintaining tractable computational complexity thanks to its sparse structure~\cite{strasdat2010why}. Such a paradigm shift has been predominantly reflected in visual or LiDAR based odometry~\cite{vins2018,RAL21_Li}. For UWB-inertial state estimation, current techniques have not fully embraced state-of-the-art methodologies, with most use cases limited to batchwise (offline) or planar motion estimation~\cite{song2021uwbimu,zheng2022uwb}.
   
    Conventional sensor fusion schemes are built atop discrete timestamps of constant interval, necessitating temporal alignment of measurements w.r.t. the estimation~\cite{vins2018}. However, different sensors fire asynchronously without any common time instant. In a single-sensor setup, measurements can also be non-uniformly sampled over time due to timestamp jitter, particularly in UWB sensing~\cite{hol2009uwb}. \red{Interpolating asynchronous range measurements from different anchors results in degraded tracking performance, especially under outliers and complex noise patterns.} Thus, continuous-time state estimation is appealing. \red{In this regard, data-driven approaches based on Gaussian processes have been systematically investigated, where various motion priors are learned within the stochastic differential equation formulations~\cite{tang2019white,barfoot2017state}}.
    
	B-splines parameterize trajectories atop \textit{knots}, or \textit{control points}, through temporal polynomials, enabling interpolation at any given time instant with locality and smoothness. This concept can be practically generalized to Lie groups or nonlinear manifolds via reformulation into a cumulative form, based on which rigid body motions can be modeled continuously over time. In~\cite{furgale2012spline,anderson2013spline}, B-splines were applied to estimating continuous-time states via batchwise maximum a posteriori. Similar offline schemes have also been proposed for attitude estimation, multi-sensor calibration, and trajectory estimation using visual/event/LiDAR-inertial setups~\cite{sommer2016continuous,furgale2013calib,patron2015spline,mueggler2018continuous,cio2022spline}. However, these batchwise schemes are computationally expensive due to the rudimentary strategies of computing time derivatives (via product rule) and Jacobians on B-splines (via numerical or automatic differentiations). A breakthrough in efficient continuous-time motion estimation using B-splines was made in~\cite{sommer2020efficient}, where recursive computation of time derivatives and Jacobians on $\SOT$ trajectories was introduced for spline fitting. This has inspired successive work in online motion estimation, with applications including RGB-D tracking, LiDAR/visual-inertial calibration and odometry~\cite{yang2021spline,javier2022spline,lv2021clins,hug2022con,persson2021practical}.
    
    While significant progress has been made in continuous-time sensor fusion, there is still a considerable gap towards more extensive engineering practice. To the best knowledge of the authors, no continuous-time solution is currently available for UWB-inertial fusion. \red{Unit quaternions are widely accepted in robotics as a nonsingular rotation representation and have certain favorable attributes in memory efficiency, numerical stability, and computation~\cite{fresk2013full,leutenegger2015keyframe,vins2018,sun2022comparative}}. However, current spline-based state estimation systems rely heavily on the theory in~\cite{sommer2020efficient} using rotation matrices, while unit quaternions are only utilized for basic arithmetic in implementation~\cite{hug2022con}. Thus, there lacks an open-source quaternion-based B-spline sensor fusion framework, with analytic kinematic interpolations and spatial differentiations unified systematically.

	\subsection*{Contribution}
	We introduce SFUISE, a novel Spline Fusion-based Ultra-wideband-Inertial State Estimation scheme (\secref{sec:sys}). Quaternion-based cubic cumulative B-splines serve as the backbone of state representation, with systematic and unified derivations of analytic kinematic interpolations and Jacobians (\secref{sec:splineEst}). Based thereon, an efficient sliding-window spline fitting scheme is established to fuse UWB-inertial readings at raw timestamps with an added option of online calibration (\secref{sec:fusion}). A dedicated study is first conducted to validate the viability of the quaternion spline fitting scheme. Afterward, we evaluate SFUISE for UWB-inertial tracking in various real-world scenarios using public data sets and experiments, including comparisons with state-of-the-art discrete-time fusion schemes (\secref{sec:eva}). The proposed system delivers real-time and superior performance over the discrete-time counterparts. Considering the generality of the proposed scheme to extensive scenarios, we open-source our implementation together with own experimental data sets.
   	\begin{figure}[t]
		\vspace{1mm}
		\centering
		\includegraphics[width=0.47\textwidth]{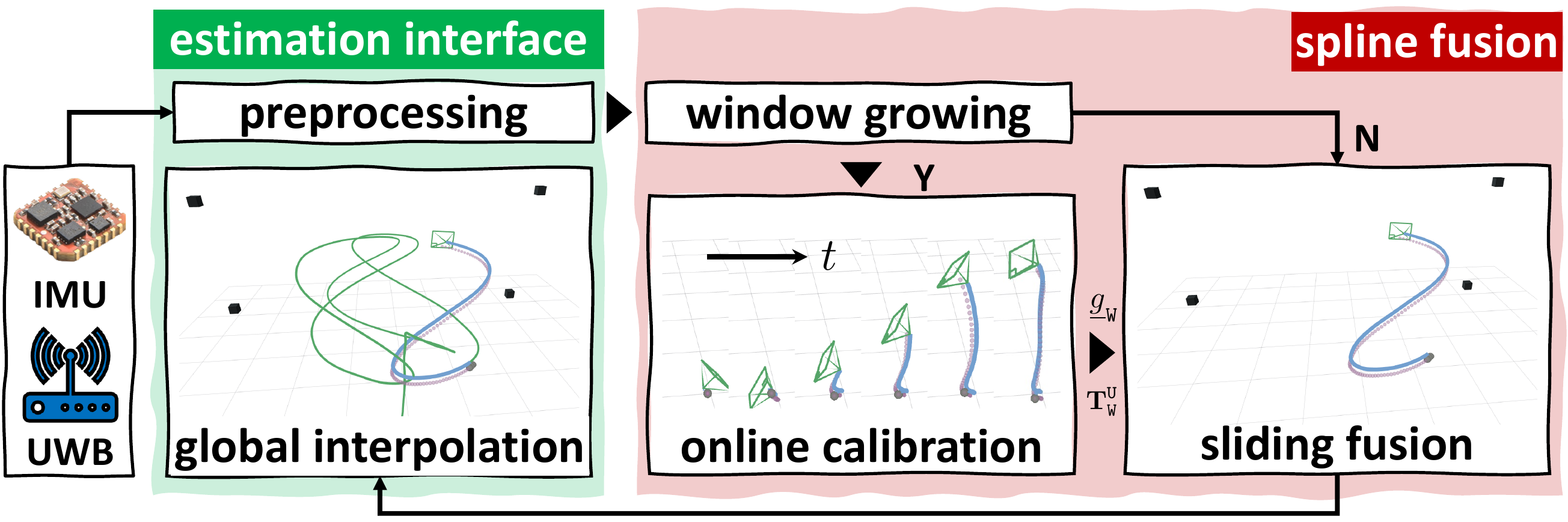}
		\caption{System pipeline of SFUISE.}
		\label{fig:sys}
	\end{figure}

 	\section{System Overview}\label{sec:sys}
	In the considered scenario, we aim to estimate the following motion-related variables
 	\begin{equation}\label{eq:state}
 		\ux(t) = [\,\ur(t)^\top,\,\us(t)^\top,\,\ub(t)^\top\,]^\top\in\Sbb^3\times\R^3\times\R^6\subset\R^{13}
 	\end{equation}
 	as a function over time. Quaternion $\ur(t)\in\Sbb^3$ and vector $\us(t)\in\R^3$ denote orientations and positions, respectively, and $\ub(t)=[\,\ub_{\acc}^\top,\ub_\gyro^\top\,]^\top\in\R^6$ incorporates biases of accelerometer and gyroscope. The proposed spline fusion-based ultra-wideband-inertial state estimation (SFUISE) system is depicted in \figref{fig:sys}. It composes an estimation interface and a functional core of spline fusion. Asynchronous UWB and IMU measurements are first preprocessed for potential downsampling, to which a cubic cumulative B-spline is fitted over a time window of limited span. As sensor data are streamed in, the spline fusion window first grows to a pre-given width, afterward slides, both w.r.t. the current timestamp for online state estimation. In the growing stage, additional online calibrations are performed to obtain the gravity vector $\ug^\fnav$ and transformation $\fT_\fnav^\fuwb$ from world ($\fnav$) to UWB ($\fuwb$) frames. Further, knot estimates given by spline fusion are sent back to estimation interface, where a global spline is maintained and interpolated for visualization.
 	
	\section{State Estimation on Cumulative B-Splines}\label{sec:splineEst}
	\subsection{Continuous-Time State Modeling}\label{subsec:spline}
	We deploy cubic (fourth-order) cumulative B-splines for continuous-time state representation. Being established upon a set of control points $\{(\cps_i,t_i)\}_{i=1}^n$ associated with time $t_i$ of uniform interval, it allows for fusing sensor readings up to the second order of motion (e.g., from accelerometer)~\cite{sommer2020efficient}. State at an arbitrary timestamp $t\in[\,t_i,t_{i+1}\,)$ can be interpolated \wrt a local set of knots $\{\cps_{i+j-2}\}_{j=0}^3$ according to
	\begin{equation}\label{eq:s}
		\us(t)=\cps_{i-2}+{\sum}_{j=1}^3\lambda_j(u)\,\ude_{j}\,,\eqwith u=\frac{t-t_i}{t_{i+1}-t_{i}}
	\end{equation}
	being the normalized time and $\ude_j=\cps_{i+j-2}-\cps_{i+j-3}$ the distance between neighboring knots. Note that we use calligraphic fonts to denote spline knots, highlighting that they do not lie on the trajectory itself. The cumulative basis functions $\{\lambda_j(u)\}_{j=1}^3$ follow $[\,\lambda_1(u),\lambda_2(u),\lambda_3(u)\,]^\top=\Phi\,\uu$\,, with
	\begin{equation*}
	\Phi={\textstyle\frac{1}{6}}\bigg[\sbbmat\,5&3&-3&1\,\\1&3&3&-2\\0&0&0&1\sebmat\bigg] \eqand \uu=[\,1,u,u^2,u^3\,]^\top
	\end{equation*}
	 endowing cubic B-splines with $\mC^2$-continuity. Expression \eqref{eq:s} can be applied to model positions and IMU biases in \eqref{eq:state}. Based on Riemannian geometry, the concept of cumulative B-splines can be naturally extended to the manifold of unit quaternions~\cite{kim1995general}. It follows
	\begin{equation}\label{eq:r}
	\ur(t)=\cpr_{i-2}\otimes{\prod}_{j=1}^{3}\Exp_{\mathds{1}}{\big(\lambda_j(u)\,\ude_j\big)}\,,
	\end{equation}
	with $\otimes$ denoting the Hamilton product and $\lambda_j(u)$ the basis functions in \eqref{eq:s}. Distance between adjacent knots is quantified via logarithm map $\ude_j=\Log_{\mathds{1}}\big({\cpr_{i+j-3}^{-1}\otimes\cpr_{i+j-2}}\big)$ at identity $\mathds{1}=[1,0,0,0]^\top$ and contributes to the on-manifold interpolation via exponential map $\Exp_\dso(\cdot)$~\cite{Li2022Dissertation}. Unless otherwise specified, the term `B-spline' in the following content refers to the cumulative formulation of uniform intervals.
	
	On modeling motion-related states, knots are optimally estimated by fitting the spline to measurements in the least squares sense. Constructing residuals in the objective refers to kinematic interpolations on cubic B-splines w.r.t. related sensory modalities. This can be computationally expensive due to large data volume and complexity in deriving motion derivatives. Meanwhile, solving nonlinear least squares requires Jacobians of on-manifold kinematic interpolations \wrt knots. In the remainder of this section, we provide these theoretical building blocks for quaternion-based B-splines towards high-performance continuous-time sensor fusion.
	
	\subsection{Kinematic Interpolations}\label{subsec:kino}
	We now present time derivatives of cubic B-splines for interpolating linear and angular velocities, and acceleration.
	\subsubsection*{Linear velocity and acceleration}
	Given a position B-spline $\us(t)$ in \eqref{eq:s}, its first-order time derivative (denoted by dot atop the variable) can be derived via $\dot{\us}(t)=\us^\prime(u){u}^\prime(t)$, with $u^\prime(t)=1/(t_{i+1}-t_i)\coloneqq1/\Dt_i$. This leads to
	\begin{equation*}
	\dot{\us}(t)={\sum}_{j=1}^3{\lambda}^\prime_j(u)\,\ude_{j}u^\prime(t)={\sum}_{j=1}^3\dot{\lambda}_j(u)\,\ude_{j}\,,
	\end{equation*}
	with $\dot{\lambda}_{j}$ being derivatives of basis functions in \eqref{eq:s} given by  $[\dot{\lambda}_1(u),\dot{\lambda}_2(u),\dot{\lambda}_3(u)]^\top=\Phi\,[\,0,1,2u,3u^2\,]^\top/\Dt_i$. Further, the acceleration over time follows
	\begin{equation}\label{eq:dds}
	\ddot{\us}(t)={\sum}_{j=1}^3\ddot{\lambda}_j(u)\,\ude_{j}\,,
	\end{equation}
	with $[\ddot{\lambda}_1(u),\ddot{\lambda}_2(u),\ddot{\lambda}_3(u)]^\top=\Phi\,[\,0,0,2,6u\,]^\top/\Dt_i^2$.
	\subsubsection*{Angular velocity}
	By definition~\cite{sola2017quaternion}, directly computing the time derivative of the quaternion spline function in \eqref{eq:r} leads to the angular velocity $\uom(t)$ in body frame via relation
	\begin{equation}\label{eq:dr}
		\dot{\ur}(t)=0.5\,\ur(t)\otimes\uom(t)\,.
	\end{equation}
	In practice, however, this requires quadratic complexity \wrt spline order due to the chain of Hamilton products. For cumulative B-splines on matrix Lie group $\SOT$, an efficient recursive method has been introduced in~\cite{sommer2020efficient}. We hereby provide the full derivation for its quaternion counterpart.
	
	For brevity, we discard the time variable in \eqref{eq:r} and split the product chain into $\ur=\ur_k\otimes\prod_{j=k+1}^{3}\Exp_{\mathds{1}}{(\lambda_j\,\ude_j)}$, where ${\ur}_k=\cpr_{i-2}\otimes\prod_{j=1}^{k}\Exp_{\mathds{1}}{(\lambda_j\,\ude_j)}$ for $k\in\{1,2,3\}$.
	The quaternion spline interpolation can then be expressed in a recursive fashion according to
	\begin{equation}\label{eq:rk}
	{\ur}_k={\ur}_{k-1}\otimes\die_k\,,
	\end{equation}
	with $\die_k\coloneqq\Exp_{\mathds{1}}{(\lambda_k\,\ude_k)}$ denoting the $k$-th spline increment obtained from exponential map. Applying the kinematic relation in \eqref{eq:dr} to $\ur_k$ yields
	\begin{equation}\label{eq:drk0}
		\dot{\ur}_k=0.5\,\ur_k\otimes\uom_k\,,
	\end{equation}
	with $\uom_k$ being the angular velocity to be computed recursively. For that, we compute the time derivative of \eqref{eq:rk} and obtain
	\begin{equation}\label{eq:drk}
		\dot{\ur}_k=\dot{\ur}_{k-1}\otimes\die_k+\ur_{k-1}\otimes\dot{\die}_k\,,
	\end{equation}
 	where time derivative of increment $\die_k$ is given by $\dot{\die}_k=\die_k\otimes(\dot{\lambda}_k\ude_k)$ according to \eqref{eq:dr}. Expression in \eqref{eq:drk} then follows
	\begin{equation*}
	\begin{aligned}
		\dot{\ur}_k&=0.5\,{\ur}_{k-1}\otimes\uom_{k-1}\otimes\die_k+\ur_{k-1}\otimes\die_k\otimes(\dot{\lambda}_k\ude_k)\\
		&=0.5\,\ur_k\otimes\big(\die_k^{-1}\otimes\uom_{k-1}\otimes\die_k+2\dot{\lambda}_k\ude_k\big)\,,
	\end{aligned}
	\end{equation*}
	based on \eqref{eq:rk} and \eqref{eq:drk0}. Subsequently, \red{the} following recursive expression for angular velocity interpolation can be established
	\begin{equation}\label{eq:omrec}
		\uom_k=\scR(\die_k^{-1})\,\uom_{k-1}+2\dot{\lambda}_k\ude_k\,,
	\end{equation}
    where $\scR(\die_k^{-1})\,\uom_{k-1}=\die_k^{-1}\otimes\uom_{k-1}\otimes\die_k$ rotates angular velocity $\uom_{k-1}$ with the inverse quaternion increment $\die_k^{-1}$. To bootstrap the recursion, we can derive $\uom_1$ via time derivative of $\ur_1=\cpr_{i-2}\otimes\die_1$, namely, $\dot{\ur}_1=\cpr_{i-2}\otimes\dot{\die}_1=\ur_1\otimes(\dot{\lambda}_1\ude_1)$, leading to $\uom_1=2\dot{\lambda}_1\ude_1$.
    
	\subsection{Spatial Differentiations}\label{subsec:jac}
	On solving the nonlinear least squares for spline fitting, the gradient of the objective can be computed via the chain rule, where the major complexity goes to the kinematic terms. For that, we consider a spline segment ranging over $t\in[\,t_i,t_{i+1})$ as introduced in \secref{subsec:spline}. Jacobians of interpolated kinematic states $\ux$ \wrt knots $\{\cpx_{i+j-2}\}_{j=0}^3$ are in principle expressed as
	\begin{equation}\label{eq:jacx}
	\frac{\dd\ux}{\dd\cpx_{i+j-2}}=\ind_{j=0}\fJ_{i-2}+\ind_{j\neq{0}}\fJ_{j}+\ind_{j\neq{3}}\fJ_{j+1}\,,
	\end{equation}
	with $\ind$ being the indicator function. Depending on the knot indices, Jacobian components in \eqref{eq:jacx} are calculated via
	\begin{equation*}
	\fJ_{i-2}\lsft{-3mu}=\lsft{-3mu}\pad{\ux}{\cpx_{i-2}}, \fJ_{j}\lsft{-3mu}=\lsft{-3mu}\pad{\ux}{\ude_j}\pad{\ude_j}{\cpx_{i+j-2}}, \fJ_{j+1}\lsft{-3mu}=\lsft{-3mu}\pad{\ux}{\ude_{j+1}}\pad{\ude_{j+1}}{\cpx_{i+j-2}}
	\end{equation*}
	that are concretized for the kinematic types as follows.
	
	\subsubsection*{Jacobians of position and orientation interpolations}
	Following \eqref{eq:jacx}, it is trivial to obtain the Jacobian of the position spline w.r.t knot as $\dd{\us}/\dd{\cps_{i+j-2}}=(\ind_{j=0}+\ind_{j\neq{0}}\lambda_j-\ind_{j\neq{3}}\lambda_{j+1})\,\eye{3}$.	For the orientation spline defined in \eqref{eq:r}, its first Jacobian component in \eqref{eq:jacx} is derived as	$\fJ_{i-2}=\lrmat{\qf}\big({\prod}_{j=1}^{3}\die_j\big)$, with function $\lrmat{\qf}$ mapping a quaternion multiplied from right-hand side into its matrix representation~\cite[Eq. 3.5]{Li2022Dissertation}. To obtain the Jacobian components $\fJ_j$ and $\fJ_{j+1}$ in \eqref{eq:jacx}, we reformulate the quaternion interpolation \eqref{eq:r} into
	\begin{equation*}
	\begin{aligned}
	  \ur&=\Big(\cpr_{i-2}\otimes{\prod}_{k=1}^{j-1}\die_k\Big)\otimes\die_j\otimes{\prod}_{k=j+1}^{3}\die_k=\llmat{\fQ}_{\la}\lrmat{\fQ}_{\ra}\,\die_j
	\end{aligned}
	\end{equation*}
	 to expose the $j$-th increment $\die_j$. $\llmat{\fQ}_{\la}$ and $\lrmat{\fQ}_{\ra}$ are matrices representing the quaternions multiplied on the left- and right-hand sides of $\die_j$ via $\llmat{\qf}$ and $\lrmat{\qf}$, respectively, i.e.,
	 \begin{equation*}
	 \llmat{\fQ}_{\la}=\llmat{\qf}\Big(\cpr_{i-2}\otimes{\prod}_{k=1}^{j-1}\die_k\Big),
	 \lrmat{\fQ}_{\ra}=\lrmat{\qf}\Big({\prod}_{k=j+1}^{3}\die_k\Big)\,.
	 \end{equation*}
	We then obtain the gradient of quaternion spline \wrt $\ude_j$ as
	\begin{equation}\label{eq:dqdd}
	\pad{\ur}{\ude_j}=\lambda_j\,\llmat{\fQ}_{\la}\,\lrmat{\fQ}_{\ra}\,\pad{\Exp_{\mathds{1}}(\uv)}{\uv}\eva{\uv=\lambda_j\ude_j}\,,
	\end{equation}
	with the gradient of exponential map provided in \eqref{eq:dexp}. The gradient w.r.t. $\ude_{j+1}$ takes the same expression as in \eqref{eq:dqdd}, which can be computed recursively given the computation result for $\ude_j$. The gradient of $\ude_j$ \wrt knot follows
	\begin{equation}\label{eq:deltaj}
	\lsft{-3mu}\pad{\ude_j}{\cpr_{i+j-2}}\lsft{-3mu}=\lsft{-3mu}\pad{\Log_{\mathds{1}}\big({\cpd_j})}{\cpr_{i+j-2}}\lsft{-3mu}=\lsft{-3mu}\pad{\Log_{\mathds{1}}(\ur)}{\ur}\eva{\ur=\cpd_j}\lsft{-25mu}\llmat{\qf}(\cpr_{i+j-3}^{-1})\,,
	\end{equation}
	with $\cpd_j\coloneqq\cpr_{i+j-3}^{-1}\otimes\cpr_{i+j-2}$. We provide the gradient of logarithm map in \eqref{eq:dlog}. The gradient of $\ude_{j+1}$ is derived as
	\begin{equation}\label{eq:gradq}
	\pad{\ude_{j+1}}{\cpr_{i+j-2}}=\pad{\Log_{\mathds{1}}(\ur)}{\ur}\eva{\ur=\cpd_{j+1}}\lrmat{\qf}\big(\cpr_{i+j-1})\,\fD\,,
	\end{equation}
	where $\fD=\diag(1,-1,-1,-1)$ denotes a diagonal matrix.
	
	\subsubsection*{Jacobian of angular velocity interpolation} In reference to \eqref{eq:jacx}, the Jacobian takes the following general form
		\begin{equation}\label{eq:omr}
		\frac{\dd{\uom}}{\dd\cpr_{i+j-2}}=\ind_{j\neq{0}}\bm\Omega_{j}+\ind_{j\neq{3}}\bm\Omega_{j+1}\,,
	\end{equation}
	with the two partial derivatives expressed as follows
	\begin{equation*}
		\bo_j=\pad{\uom}{\uom_j}\pad{\uom_j}{\ude_j}\pad{\ude_j}{\cpr_{i+j-2}}\,,	\bo_{j+1}=\pad{\uom}{\uom_{j+1}}\pad{\uom_{j+1}}{\ude_{j+1}}\pad{\ude_{j+1}}{\cpr_{i+j-2}}\,.
	\end{equation*}
	The two components above are computed in the same fashion, and we now only demonstrate the derivation for $\bo_j$. The first term in $\bo_j$ can be obtained via the recursion in \eqref{eq:omrec} according to the following derivation. For cubic cumulative B-spline, we have $\uom(t)=\uom_{k}(t)$ with $k=3$. Thus, we obtain 
	\begin{equation*}
		\pad{\uom}{\uom_j}=\pad{\uom}{\uom_3}{\prod}_{k=1}^{3-j}\pad{\uom_{4-k}}{\uom_{3-k}}={\prod}_{k=1}^{3-j}\scR(\die_{4-k}^{-1})\,,
	\end{equation*}
	and the second term is derived as
	\begin{equation*}
		\pad{\uom_j}{\ude_j}=\lambda_j\pad{(\scR(\ur)\,\uom_{j-1})}{\ur}\eva{\ur=\die_j^{-1}}\lsft{-10mu}\fD\pad{\Exp_\dso(\uv)}{\uv}\eva{\uv=\lambda_j\ude_j}\lsft{-10mu}+2\dot{\lambda}_j\eye{3}\,.
	\end{equation*}
	$\lpad{(\scR(\ur)\uom_{j-1})}{\ur}$ denotes the Jacobian of quaternion rotation that is given by in~\cite[Eq.19]{sola2017quaternion}. And the last term $\lpad{\ude_j}{\cpr_{i+j-2}}$ is available in \eqref{eq:deltaj}.

	\subsubsection*{Jacobian of acceleration interpolation}
	We apply the general formulation \eqref{eq:jacx} to the acceleration interpolation \eqref{eq:dds}. It is then straightforward to obtain the Jacobian as follows
	\begin{equation*}
	\fad{\ddot{\us}}{\cps_{i+j-2}}=(\ind_{j\neq{0}}\ddot{\lambda}_j-\ind_{j\neq{3}}\ddot{\lambda}_{j+1})\eye{3}\,.
	\end{equation*}

	\section{Online UWB-Inertial Spline Fusion}\label{sec:fusion}
	\begin{figure}[t]
		\vspace{1mm}
		\centering
		\includegraphics[width=0.485\textwidth]{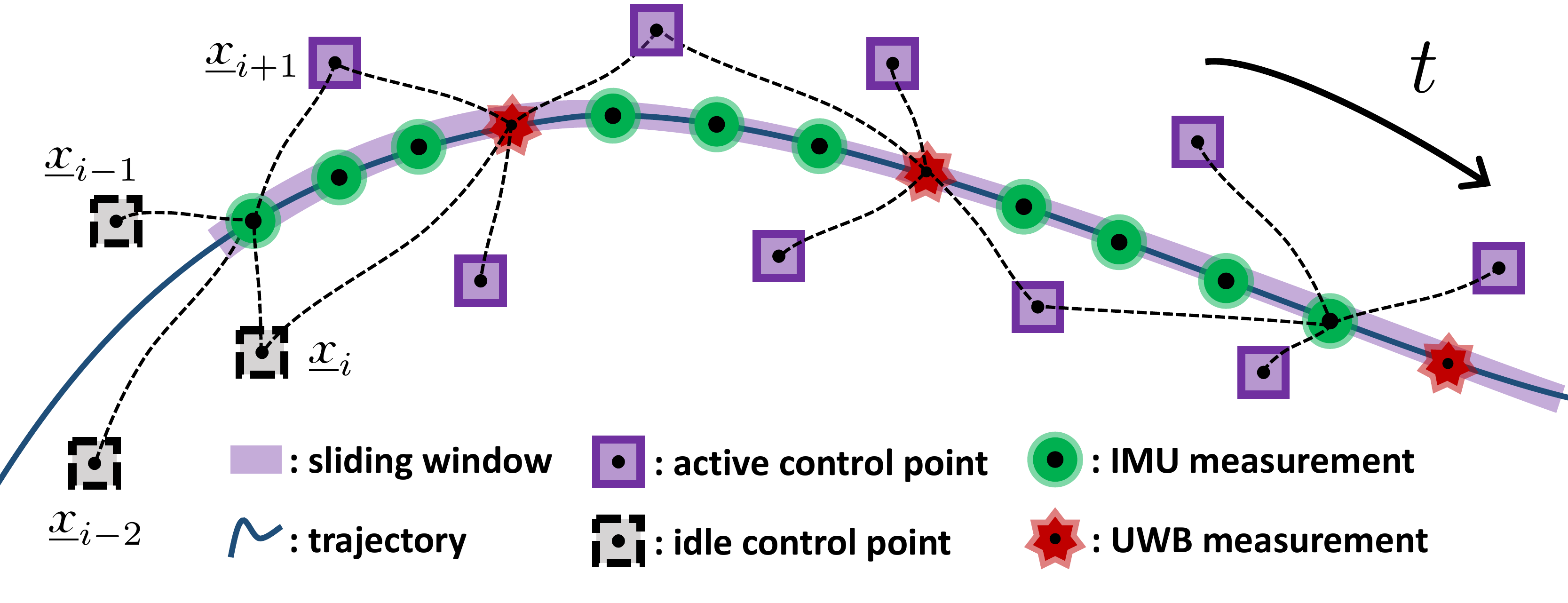}		
		\caption{Sliding-window spline fitting for online estimation.}
		\label{fig:spline}
	\end{figure}
	\subsection{Sliding-Window Spline Fitting}\label{subsec:window}
	 Shown in \figref{fig:spline}, we exploit cubic B-splines to parameterize state \eqref{eq:state} continuously over time with knots concatenated as $\cpx=[\,\cpr^\top,\cps^\top,\cpb^\top]^\top\in\R^{13}$. To keep the computational intensity tractable for online performance, we bound the spline fitting problem over a window of recent $\tau_\twi$ knots 	$\cpX_\twi=[\,\cpx_1,...,\cpx_{\tau_\twi}\,]\in\R^{13\times\tau_\twi}$, namely,
	\begin{equation}\label{eq:opt}
	 \cpX_\twi^\star=\textstyle\argx{\min}{\cpX_\twi}\scF(\cpX_\twi)\,,
	\end{equation}
	with the objective function formulated as $\scF(\cpX_\twi)=$
	\begin{equation}\label{eq:obj}
	\scv_\uwb\sum_{i=1}^{m}\big\Vert\scE_\uwb(\cpX_\twi,\hzz_{\uwb,i})\big\Vert^2_{\fC_\uwb}+\scv_{\imu}\sum_{k=1}^{n}\big\Vert\scE_\imu(\cpX_\twi,\huz_{\imu,k})\big\Vert^2_{\fC_\imu}\,.
	\end{equation}
	$\scE_\uwb$ and $\scE_\imu$ denote residual terms built upon UWB and IMU measurements, $\{\hzz_{\uwb,i}\}_{i=1}^m$ and $\{\huz_{\imu,k}\}_{k=1}^n$, respectively, each observed at raw timestamps. \red{$\fC_\circ$ and $\scv_\circ$ indicate the sensor noise covariances} and tunable weights, respectively, for each error term (subscript '$\circ$' stands for '$\uwb$' or '$\imu$' denoting UWB or IMU, respectively). As the window slides, the so-called \textit{active} knots within the current window are updated over time until to be dropped out. Meanwhile, the three knots that are most recently removed from the window are turned into an \textit{idle} state. They are no longer being optimized, however, still participate in computing residuals in the current window via kinematic interpolations, which enables motion continuation across sliding windows over time~\cite{lv2021clins}. Note that timestamps of residuals and underlying knots are not to be aligned. This allows for flexible multi-sensor fusion and efficient motion representation compared with discrete-time paradigm. 
	
	We now specify the residual terms in \eqref{eq:obj}. Throughout the following derivations, we use B-splines to describe the $6$-DoF motion of the IMU body (\fimu) \wrt the world frame ($\fnav$). Anchor positions are given \wrt UWB frame ($\fuwb$) with a transformation $\fT_\fnav^\fuwb\in\SET$ from the world frame.
	
	\subsubsection*{UWB residual}
	We demonstrate the UWB residual in \eqref{eq:obj} for time-of-arrival (ToA) ranging. Given a range measurement $\hzz_{\uwb,i}$ at timestamp $t_i$, its residual term follows
	\begin{equation}\label{eq:ruwb}
	\scE_\uwb(\cpX_\twi,\hzz_{\uwb,i})=\big\Vert{\fT_\fnav^\fuwb\,\us_{\tta,i}^\fnav-\ual_{\ttn,i}^\fuwb}\big\Vert-\hzz_{\uwb,i}\,,
	\end{equation}
	with $\us_{\tta,i}^\fnav=\ur(t_i)\otimes\unu_\tta^\fimu\otimes\ur^{-1}(t_i)+\us(t_i)$ being the UWB tag position in world frame obtained by transforming tag coordinates $\unu_\tta^\fimu$ \wrt body pose interpolated at $t_i$. $\ual_{\ttn,i}^\fuwb$ denotes coordinates of the corresponding anchor. Residual for time-difference-of-arrival (TDoA) ranging can be obtained similarly, which we do not specify due to page limit.
	 	 
	\subsubsection*{IMU residual}
	Given the $k$-th IMU reading $\huz_{\imu,k}$ of acceleration $\hua_k^\fimu$ and angular velocity $\huo_k^\fimu$, we construct residual $\scE_\imu(\cpX_\twi,\huz_{\imu,k})=[\,\ez_{\acc,k}^\top,\,\ez_{\gyro,k}^\top,\ez_{\bias,k}^\top\,]^\top$
	for spline fitting to inertial data. The accelerometer residual $\ez_{\acc,k}$ is
	\begin{equation}\label{eq:racc}
		\ez_{\acc,k}=\ua^\fimu_k+\ub_{\acc,k}-\hua_k^\fimu\,,
	\end{equation}
	with $\ua^\fimu_k=\ur^{-1}(t_k)\otimes\big(\ddot{\us}(t_k)+\ug^\fnav\big)\otimes\ur(t_k)$ transforming the interpolated acceleration $\ddot{\us}(t_k)$ together with gravity $\ug^\fnav$ to body frame. $\ub_{\acc,k}$ is the accelerometer bias interpolated on spline $\ub(t)$ at $t_k$. The gyroscope residual $\ez_{\gyro,k}$ follows $\ez_{\gyro,k}=\uom^\fimu(t_k)+\ub_{\gyro,k}-\huo_k^\fimu$,	with $\uom^\fimu(t_k)$ interpolated recursively as shown in \eqref{eq:omrec}. $\ub_{\gyro,k}$ denotes the gyroscope bias interpolated at $t_k$. Further, we compute the difference of consecutive bias interpolations at $t_k$ and $t_{k+1}$ as the IMU bias residual, namely, $\ez_{\bias,k}=\ub(t_{k+1})-\ub(t_k)$.

	\subsection{Concurrent Calibration}
	In general, the transformation $\fT_\fnav^\fuwb$ (parameterized by a quaternion $\uq_\fnav^\fuwb\in\Sbb^3$ and a translation vector $\ut_\fnav^\fuwb\in\R^3$) from world to UWB frames in \eqref{eq:ruwb} is unknown and typically dependent on the anchor coordinates and \red{the} tag pose at system initialization. Also, the gravity orientation $\bar{\ug}^\fnav$ (obtained via normalizing the gravity $\ug^\fnav=\nm{\ug^\fnav}\,\bar{\ug}^\fnav$) in \eqref{eq:racc} is in general not available upon navigation. To approximate it, a common practice is to average the first several accelerometer readings under the assumption of a static motion at start. This can be easily violated by an undesirable starting condition (e.g., on the fly). To address these issues, we concatenate $(\uq_\fnav^\fuwb,\ut_\fnav^\fuwb)$ and $\bar{\ug}^\fnav$ after the knots in the state vector \eqref{eq:opt} during window-growing phase and estimate them via spline fitting.
 	\begin{figure*}[t]
		\vspace{1mm}
		\centering
		\begin{tabular}{ccccc}
			&{\textbf{\ttt{const1}}} &\textbf{\ttt{const2}} &\textbf{\ttt{const3}} &\textbf{\ttt{const4}}\\
			\toprule
			\multirow[t]{1}{*}{\hspace{-1mm}\rotatebox{90}{~~~\textbf{\ttt{tdoa2}}}\hspace{-6mm}}&
			\adjustbox{trim={0.005\width} {0.025\height} {0.09\width} {0.02\height},clip}{\includegraphics[width=0.150\textwidth]{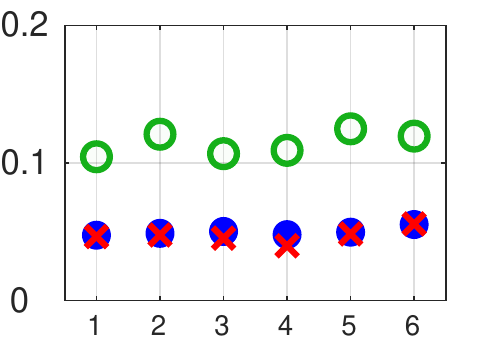}}&
			\adjustbox{trim={0.005\width} {0.025\height} {0.09\width} {0.02\height},clip}{\includegraphics[width=0.150\textwidth]{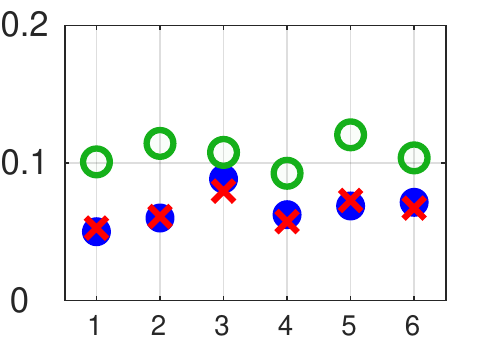}}&
			\adjustbox{trim={0.035\width} {0.025\height} {0.09\width} {0.02\height},clip}{\includegraphics[width=0.200\textwidth]{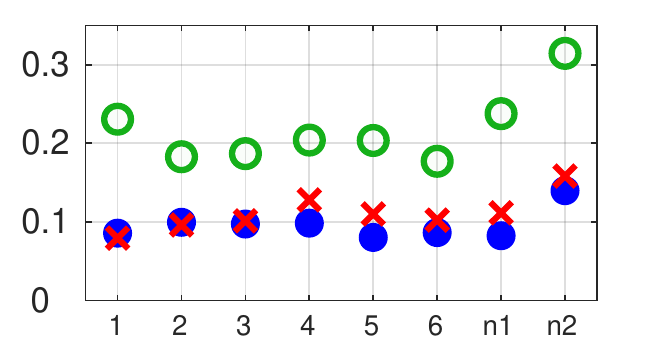}}&
			\adjustbox{trim={0.090\width} {0.025\height} {0.09\width} {0.01\height},clip}{\includegraphics[width=0.535\textwidth]{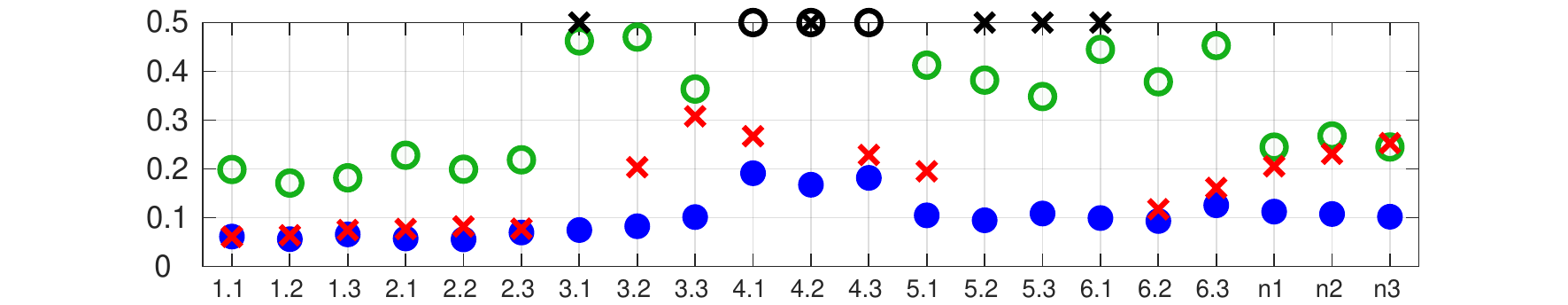}}\\
			\midrule
			\multirow[t]{1}{*}{\hspace{-1mm}\rotatebox{90}{~~~\textbf{\ttt{tdoa3}}}\hspace{-6mm}}&
			\adjustbox{trim={0.005\width} {0.025\height} {0.09\width} {0.02\height},clip}{\includegraphics[width=0.150\textwidth]{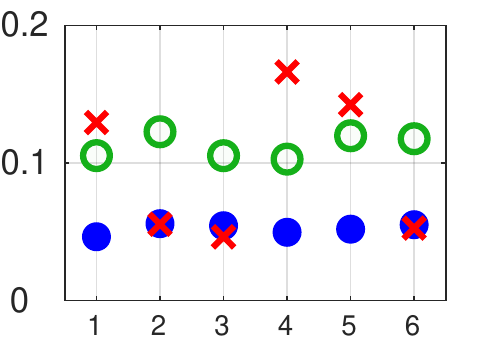}}&
			\adjustbox{trim={0.005\width} {0.025\height} {0.09\width} {0.02\height},clip}{\includegraphics[width=0.150\textwidth]{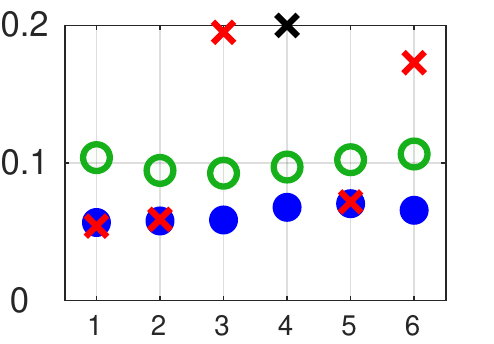}}&
			\adjustbox{trim={0.035\width} {0.025\height} {0.09\width} {0.02\height},clip}{\includegraphics[width=0.200\textwidth]{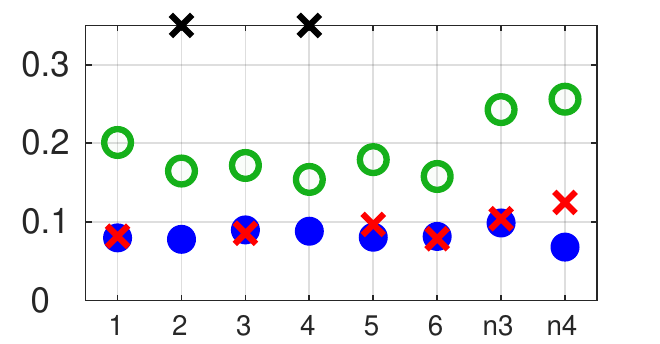}}&
			\adjustbox{trim={0.090\width} {0.025\height} {0.09\width} {0.01\height},clip}{\includegraphics[width=0.535\textwidth]{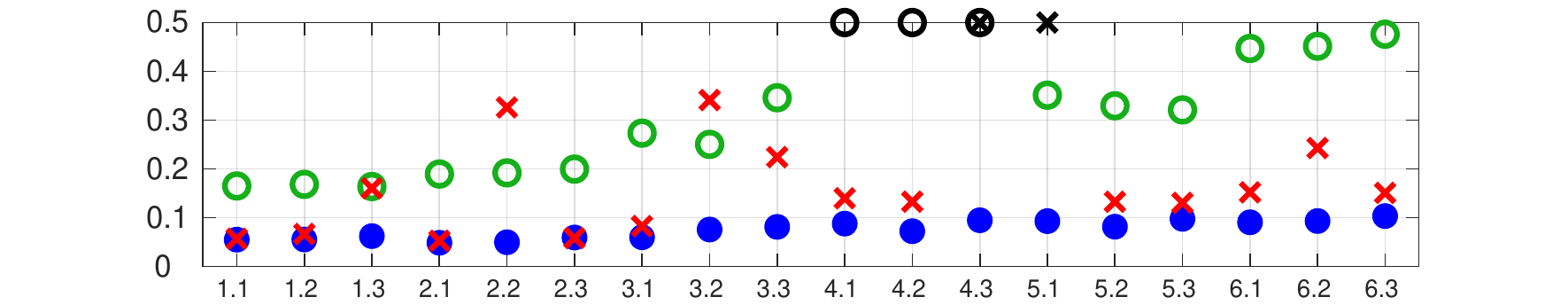}}\\
			\bottomrule
		\end{tabular}
		\caption{APEs obtained from benchmark on all sequences of \ttt{UTIL} data set. The vertical axes denote RMSEs in meters. The horizontal axes denote \ttt{trial\#} in \ttt{const1} to \ttt{const3} and \ttt{trial\#}. \ttt{traj\#} in \ttt{const4}, where \ttt{n} indicates a \ttt{manual} sequence. `\ttt{\#}' denotes the sequence index. Results from SFUISE are plotted with {\color{blue}$\bullet$}. Results from ESKF and GUIF are given by {\color{mGreen}{$\pmb{\pmb{\circ}}$}} and {\color{Red}$\pmb{\pmb\times}$}, with $\boldsymbol\circ$ and $\boldsymbol\times$ indicating tracking fails, respectively.}
		\label{fig:util}
	\end{figure*}

	\subsection{Implementation}
	The proposed spline fusion-based UWB-inertial state estimation (SFUISE) scheme is developed in C++ using ROS~\cite{quigley2009ros}. The system composes two individual nodes corresponding to the two functional blocks in the system pipeline \figref{fig:sys}. Besides basic computational tools such as Eigen (https://eigen.tuxfamily.org), no further external dependency is required. We customize Levenberg-Marquardt (LM) algorithm to solve the nonlinear least squares in \eqref{eq:opt} iteratively. The closed-form gradients w.r.t. quaternion states are further established in the tangent space at estimate $\check{\cpr}$ w.r.t. a local perturbation $\uph\in\R^3$ by multiplying with $\pad{({\check{\cpr}}\otimes\Exp_\dso({\uph}))}{\uph}\big\vert_{\uph=\uzero}=\llmat{\qf}(\check{\cpr})\pad{\Exp_{\dso}(\uph)}{\uph}\big\vert_{\uph=\uzero}$\,. Solving the linearized system yields an increment $\uph^*$, which updates the estimate through $\check{\cpr}\gets\check{\cpr}\otimes\Exp_{\dso}(\uph^*)$~\cite{grisetti2010tutorial,leutenegger2015keyframe}. Furthermore, the gravitation orientation $\bar{\ug}^\fnav\in\Sbb^2$ is handled in a similar fashion as introduced in~\cite{vins2018}.
	
	\section{Evaluation}\label{sec:eva}
	In this section, we first provide a rigorous validation for the proposed quaternion spline fitting scheme. Afterward, we conduct in-depth benchmarks of SFUISE in diverse real-world scenarios. All evaluations are conducted using a laptop (Intel i7-12800H CPU, $32$ GB RAM) running Ubuntu 20.04.
	\subsection{Validation of Quaternion Spline Fitting}\label{sec:validation}
	The major theoretical complexity for the proposed spline fusion schemes lies in the part for orientation. Thus, we adopt the same synthesis as presented \red{in~\cite[Sec.~6.1]{sommer2020efficient}} for batchwise continuous-time $\SOT$ estimation using both orientation and angular velocity measurements. We compare our quaternion-based scheme with the one using rotation matrices in \cite{sommer2020efficient} for fitting cubic cumulative B-splines. Both schemes are equipped with the same objective and stopping criteria. For each scheme, we exploit the Ceres Solver (http://ceres-solver.org) using LM algorithm with auto-differentiation for gradient computation, and the custom LM solvers using corresponding analytic gradients. In addition, we integrate our quaternion-based  analytic gradient into Ceres for validation. We scale up the number of knots and measurements in the original sequence by a factor of $\{1,5,10\}$ and compute average runtime over $500$ runs with knot initializations around identities perturbed by a small random noise.  All methods have converged with an averaged RMSE of $\sn{2.21}{-4}$  w.r.t. the ground truth in terms of the $\SOT$ metric in~\cite[Eq. 19]{huynh2009metrics}. As shown in Tab.~\ref{tab:valid}, the proposed quaternion spline fitting scheme achieves superior runtime efficiency, with gradients obtained from both auto-differentiation and analytic expressions. By utilizing closed-form gradients, our quaternion-based scheme runs slightly faster within Ceres than the custom LM in~\cite{sommer2020efficient}, which employs splines defined on rotation matrices.
	\begin{table}[h]
		\centering
		\begin{tabular}{r|cc|ccc}
			\toprule
			&\multicolumn{2}{c|}{Auto. Diff.*}  &\multicolumn{3}{c}{Analytic} \\
		  	\midrule
			\ttt{Factor}  &\cite{sommer2020efficient} &Ours &\cite{sommer2020efficient}  &Ours* &Ours\\
			\midrule
			\ttt{$\times$1}  &$0.0365$ &$\bf0.0274$ &$0.0201$ &$0.0184$ &$\bf0.0137$\\
			\ttt{$\times$5}  &$0.2161$ &$\bf0.1748$ &$0.1198$ &$0.1157$ &$\bf0.0833$\\
			\ttt{$\times$10}  &$0.4983$ &$\bf0.4083$ &$0.2711$ &$0.2660$ &$\bf0.1850$\\
			\bottomrule
		\end{tabular}
	\caption{Comparisons of different schemes of orientation spline fitting w.r.t. runtime in seconds. Methods with `*' exploit Ceres for optimization. Otherwise, the custom solvers are implemented using the same LM algorithm. The best runtime in each category is highlighted in bold.}
	\label{tab:valid}
	\vspace{-5mm}
	\end{table}
	
	\subsection{Benchmarking Setup}\label{subsec:setup}
	 In the upcoming sections, the proposed system is evaluated in real-world scenarios based on the public data set \texttt{UTIL}~\cite{zhao2022util} and own experiments, incorporating both ToA and TDoA ultra-wideband data. Two major discrete-time sensor fusion schemes from the state of the art are considered for comparison. An own composition of graph-based UWB-inertial fusion (GUIF) system is developed with reference to \cite{giovanni2020tight} and \cite{RAL21_Li}.  Here, discrete-time states are estimated via sliding window optimization with residuals of IMU preintegration, UWB ranging, and the prior factor from marginalization. Online calibration of $\fT_\fnav^\fuwb$ is enabled during window-growing stage. Further, we deploy the error-state Kalman filter (ESKF) provided by~\cite{zhao2022util} with default calibration parameters for evaluation against the recursive estimation scheme. In order to achieve functional UWB ranging under signal interference, the three systems are equipped with a simple thresholding step to reject outliers in UWB measurements. Throughout the benchmark, SFUISE is configured with a sliding window of $100$ knots at $10$ Hz. All UWB readings are exploited for state estimation without downsampling. In order to devote the focus to investigating the core sensor fusion scheme, we avoid fine-tuning of system configuration parameters.

	\subsection{Public Data Set}
	We exploit \ttt{UTIL} flight data set for evaluating the proposed UWB-inertial fusion scheme with TDoA ultra-wideband ranging. Overall $79$ sequences are collected onboard a quadcopter mounted with a UWB tag and an IMU ($1000$ Hz). The UWB sensor network is low-cost and operated at both centralized (\ttt{tdoa2}) and decentralized (\ttt{tdoa3}) modes under four different anchor constellations (\ttt{const1-4}). The TDoA  measurements are collected with data rates ranging from $200$ to $500$ Hz including numerous challenging scenarios created by static and dynamic obstacles of various types of materials. Some sequences also exhibit an absence of IMU measurements. Given knot estimates from SFUISE, we obtain the global pose trajectory via interpolation at the frame rate of ground truth ($200$ Hz), \wrt which we compute the RMSE of the absolute position error (APE) to quantify the tracking accuracy.
	
\begin{figure}[t!]
		\centering
		\begin{tabular}{ccc}
			\adjustbox{trim={0.35\width} {0.32\height} {0.3\width} {.35\height},clip}{\includegraphics[width=0.4\textwidth]{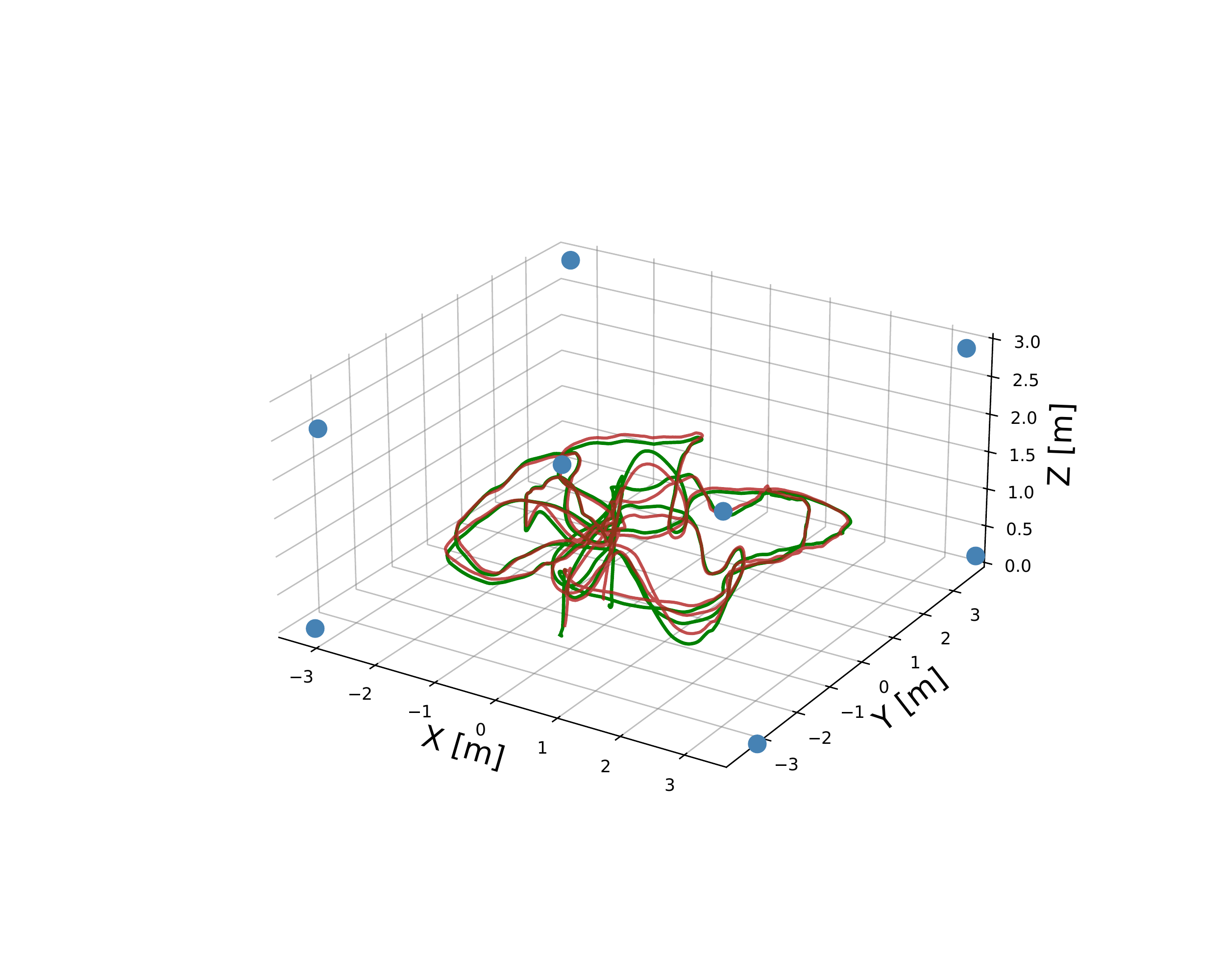}}&
			\adjustbox{trim={0.35\width} {0.32\height} {0.3\width} {.35\height},clip}{\includegraphics[width=0.4\textwidth]{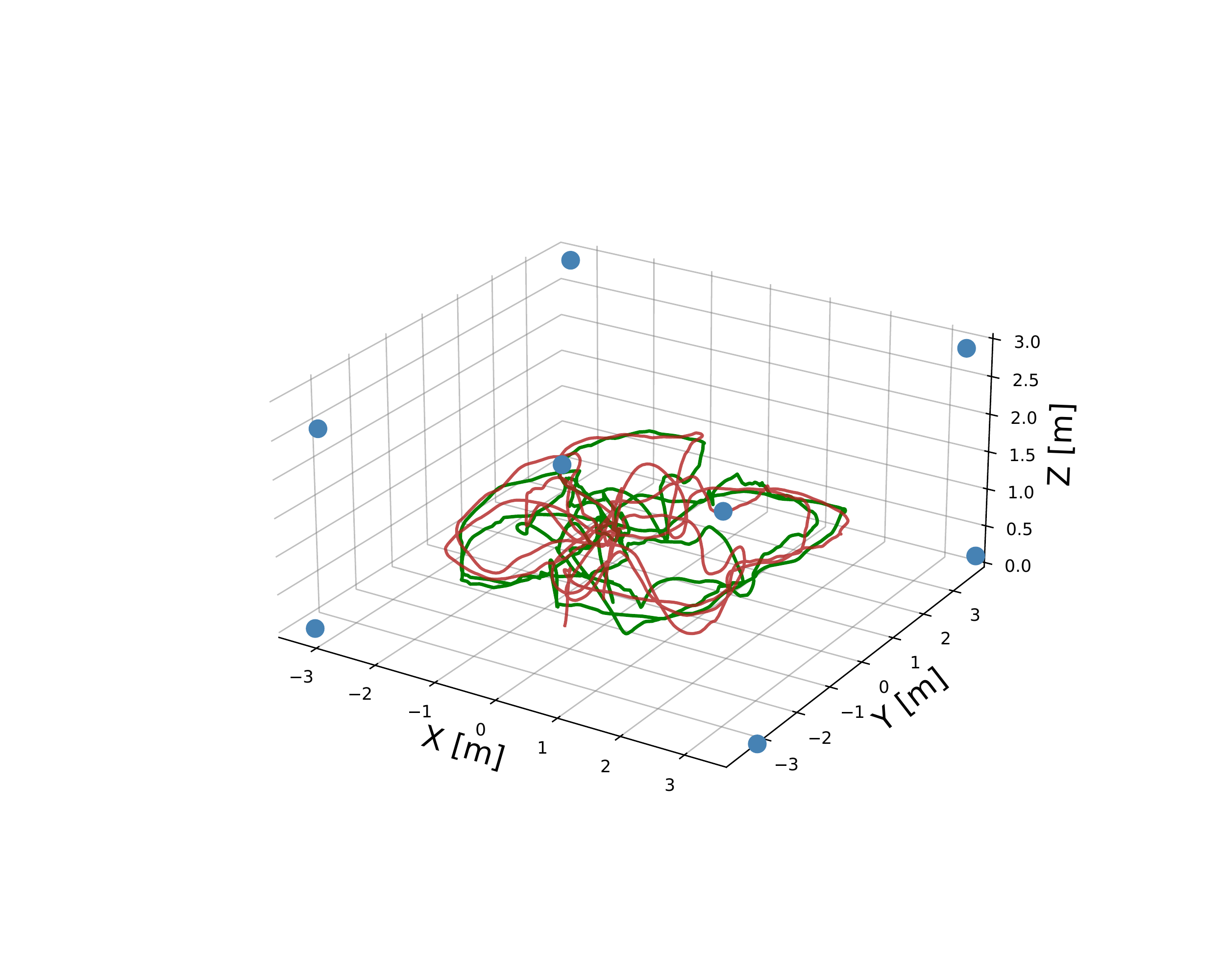}}&
			\adjustbox{trim={0.35\width} {0.32\height} {0.3\width} {.35\height},clip}{\includegraphics[width=0.4\textwidth]{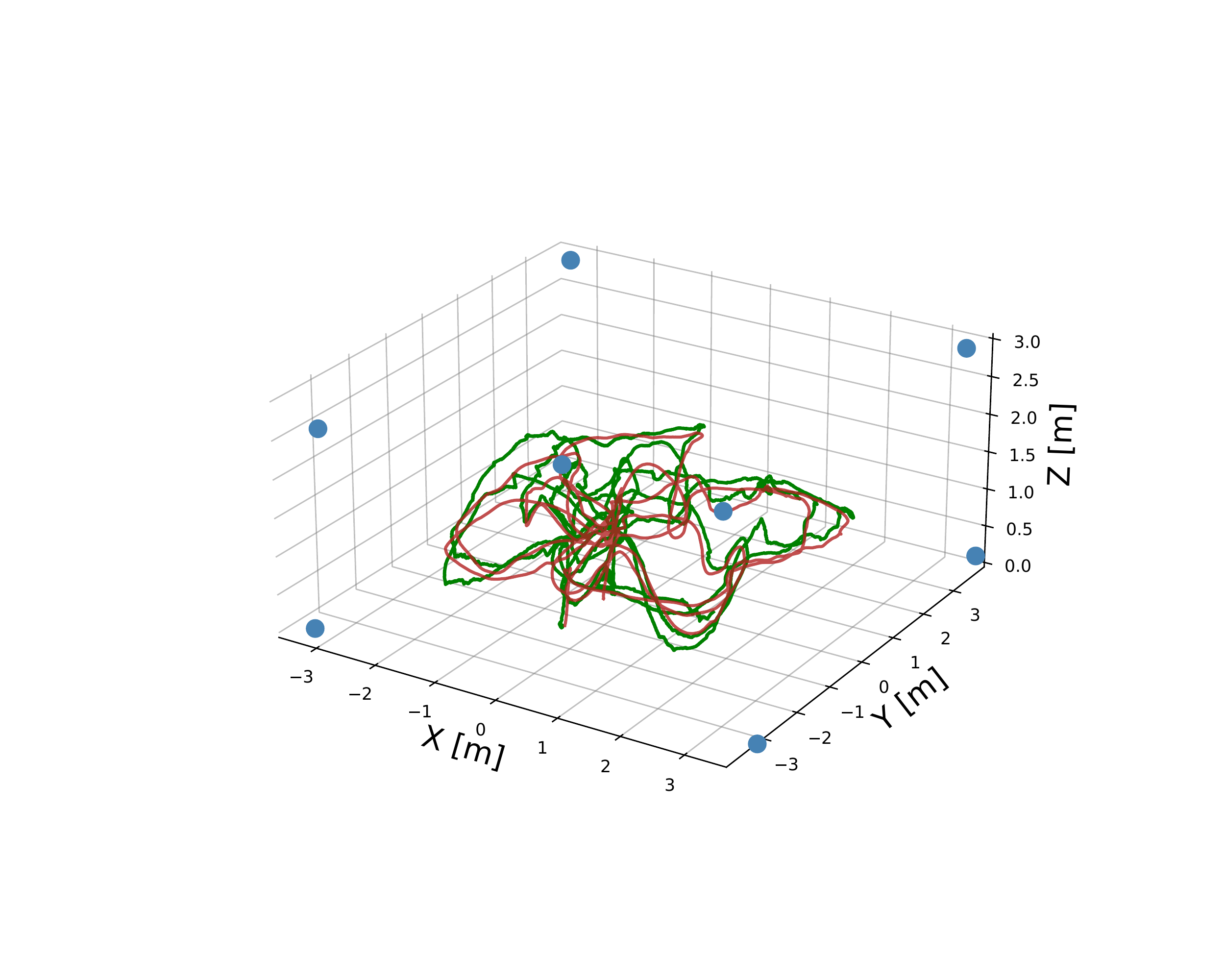}}\\
			\subcap{SFUISE} & \subcap{GUIF} & \subcap{ESKF}
		\end{tabular}
		\caption{A qualitative comparison of evaluated methods on \ttt{const4-trial7-tdoa2-manual1}. Ground truth and estimates are depicted in red and green, respectively.}
		\label{fig:compare}
		\vspace{-4mm}
	\end{figure}

	As shown in \figref{fig:util}, results given by the three systems are summarized w.r.t. UWB operation modes and anchor constellations. The proposed spline fusion scheme delivers superior performance over discrete-time approaches using recursive estimation or graph-based optimization. In particular, it exhibits a fairly good robustness against challenging conditions, e.g., in sequences with \ttt{const4}, where the tracking space is cluttered with static and dynamic obstacles of different materials including metal. In the face of time-varying NLOS and multi-path interference, discrete-time approaches are more susceptible to these effects, especially during online calibration, leading to large tracking errors or even complete failure. For demonstration, we plot results of a few representative runs of SFUISE in \figref{fig:sfuise}. Another qualitative comparison of spline fusion with discrete-time schemes is shown in \figref{fig:compare} based on a representative sequence.

	\subsubsection*{Runtime}
	As shown in \figref{fig:sys}, the two functional modules in the system pipeline run in parallel, and the spline fusion module dominates the computational cost compared with the lightweight estimation interface. Therefore, we record runtime of the backend fusion module \wrt the frame rate of knots at $10$ Hz (thus $100$ ms available for computation in real time). The proposed system delivers real-time performance with average runtime of $42.3\pm1.9$ ms and $37.2\pm1.8$ ms per sliding step throughout subsets \ttt{tdoa2} and \ttt{tdoa3}, respectively. The small standard deviations indicate a well-bounded computational cost during sliding-window spline fusion. According to our investigation, solving the nonlinear least squares in \eqref{eq:obj} using our custom LM method usually converges within five iterations.

	\subsection{Experiment}
\begin{figure}[t!]
		\vspace{1mm}
		\centering
		\begin{minipage}[b]{0.5\textwidth}
			\begin{minipage}[b]{0.3\textwidth}
				\centering
				\includegraphics[width=0.97\textwidth]{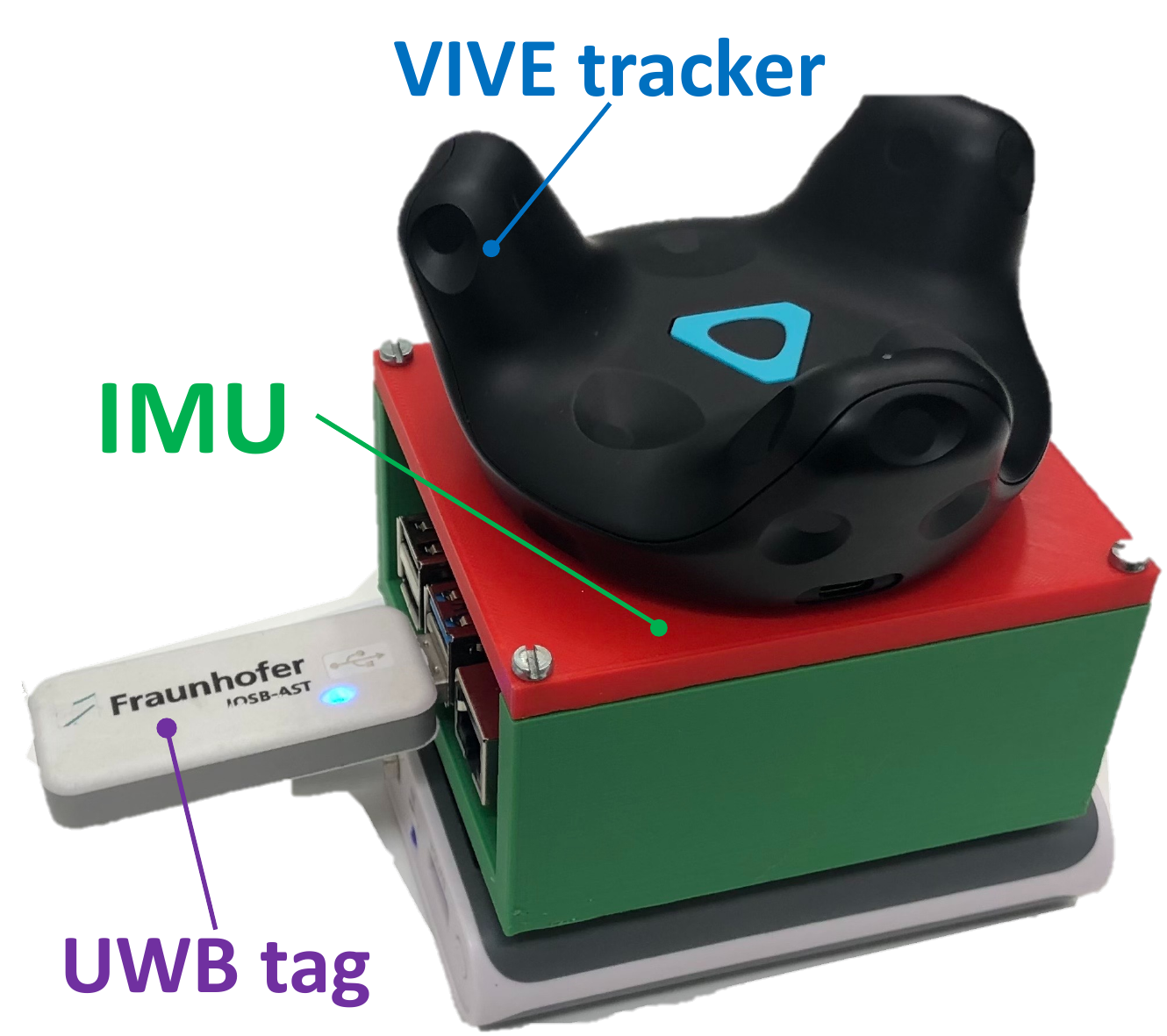}
				\captionof*{figure}{\subcap{(A) sensor suite}}
			\end{minipage}
			\hspace{-2mm}
			\begin{minipage}[b]{0.5\textwidth}
				\centering
				\begin{tabular}{cccc}
					\toprule
					{Sequence} &{SFUISE} &{GUIF}	&{ESKF} \\ 
					\midrule
					{\ttt{Walk1}} & {$0.117$} & {$0.143$} & {$0.225$}\\
					{\ttt{Walk2}} & {$0.094$} & {$0.098$} & {$0.245$}\\
					{\ttt{Walk3}} & {$0.094$} & {$0.229$} & {$0.270$}\\
					\bottomrule
				\end{tabular}
				\captionof*{table}{\subcap{(B) APE (RMSE) in meters}.}
			\end{minipage}
		\end{minipage}
		\caption{Miniature sensor suite for recording \ttt{ISAS-Walk} and corresponding APEs as shown in (A) and (B), respectively.}
		\label{fig:isas}
		\vspace{-2mm}
	\end{figure}

	To further evaluate the proposed scheme on UWB-inertial fusion using ToA ranging, a miniature sensor suite has been instrumented as shown in \figref{fig:isas}-(A). It is composed of a UWB tag provided by Fraunhofer IOSB-AST and an IMU embedded on Sense HAT (B), both mounted to a Raspberry Pi (https://www.waveshare.com) for sensor coordination and data recording. An additional VIVE tracker (https://www.vive.com) is added to provide the ground truth. Overall three sequences, \ttt{ISAS-Walk1}, \ttt{ISAS-Walk2} and \ttt{ISAS-Walk3}, are recorded during indoor walks with inertial and ultra-wideband (including five anchors) readings both at a frame rate of about $80$ Hz. We list the APE (RMSE) in meters given by the three systems in \figref{fig:isas}-(B). The proposed scheme SFUISE produces the best tracking accuracy consistently throughout the data set. 

	\subsection{Discussion}\label{subsec:discuss}
\begin{figure}[t!]
		\vspace{1mm}
		\centering
		\begin{tabular}{cc}
			\includegraphics[width=0.2\textwidth]{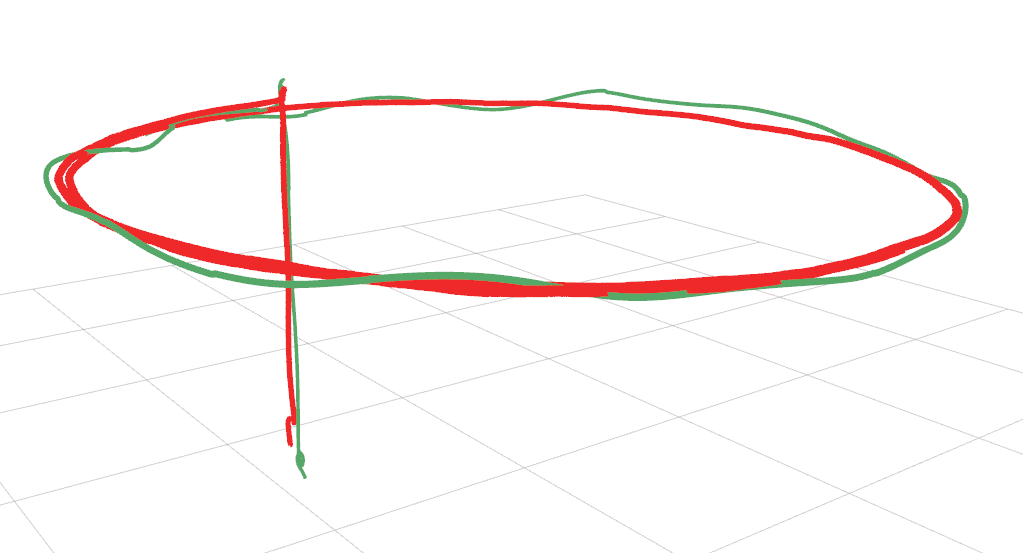} &
			\includegraphics[width=0.2\textwidth]{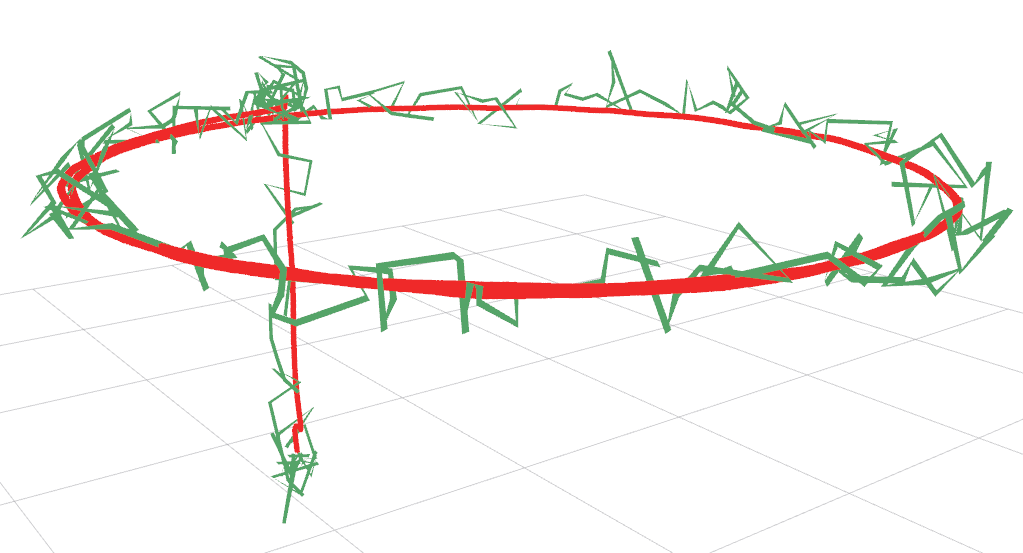}\\
			\subcap{(A) spline-based} &\subcap{(B) graph-based}
		\end{tabular}
		\caption{UWB-only tracking using SFUISE (A) and GUIF (B). Estimates  (green) are depicted w.r.t. ground truth (red) at $200$\,Hz. The spline-based approach guarantees physically feasible estimates inherently.}
		\label{fig:uwbonly}
		\vspace{-2mm}
	\end{figure}

	As verified in the mass evaluation based on \ttt{UTIL} and \ttt{ISAS-Walk}, SFUISE shows superior performance over discrete-time fusion schemes, particularly, in unfavorable scenarios of signal interference and complex noise patterns that are hard or infeasible to detect and model. Since any kinematic interpolation refers to four consecutive knots (for cubic B-splines), the induced pose trajectory inherently guarantees motion continuation up to the second order. In comparison with discrete-time fusions, this implicitly brings extra constraints to solving the nonlinear optimization problem in spline fitting, while still acknowledging the locality of observations. As a result, sensor fusion exhibits better stability and robustness under challenging conditions even together with online calibration. To further highlight the strength of spline fusion in motion estimation, we select \ttt{const1-trial1-tdoa2} of \ttt{UTIL}, and discard the IMU readings for UWB-only navigation. We run SFUISE and GUIF online at estimation frame rates of $1$ Hz and $10$ Hz, respectively, with sliding windows both configured as $10$ seconds to guarantee same amount of TDoA measurements. No explicit kinematic constraint, such as motion smoothness, is involved into state estimation. A qualitative comparison is demonstrated in \figref{fig:uwbonly}. Due to signal noise and interference, graph-based approach produces physically-infeasible motion estimates. The proposed spline-based scheme delivers a smooth trajectory with an efficient state representation using knots estimated at $1$\,Hz ($1/10$ memory consumption of the discrete-time states delivered at $10$\,Hz by GUIF).
	
	\section{Conclusion}\label{sec:conc}
	In this work, we propose a new framework for continuous-time state estimation using ultra-wideband-inertial sensors. Quaternion-based cubic B-splines are exploited for six-DoF motion representation, based on which we systematically derive a unified set of theoretical tools for efficient spline fitting, including analytic kinematic interpolations and spatial differentiations on B-splines. This further facilitates the establishment of the novel sliding-window spline fusion scheme for online UWB-inertial state estimation. The resulted system, SFUISE, is evaluated in real-world scenarios based on public data set and experiments. It is real-time capable and shows superior performance over major discrete-time estimation approaches \wrt tracking accuracy, robustness and deployment flexibility. There still remains considerable potential to exploit the proposed scheme. B-splines of nonuniform knots can be further investigated to achieve more efficient state representation and estimation. Based on our theoretical contribution and open-source implementation, extensive engineering practice can be equipped with continuous-time paradigm in areas of automatic control, path \red{planning}, odometry and mapping, etc. 
	
	\section*{Appendix}
	\subsection{Jacobian of quaternion exponential map}
	For any point $\uv\in\R^3$ in the tangent space at identity $\dso$ on the manifold of unit quaternions, it can be retracted to $\Sbb^3$ via exponential map according to 
	\begin{equation*}
		\Exp_\dso(\uv)=[\,\cos\nm{\uv},\uv^\top\sinc\nm{\uv}\,]^\top\in\Sbb^3\,,
	\end{equation*}
    with $\nm{\cdot}$ denoting the $\mathscr{L}_2$ norm~\cite{Li2022Dissertation}. The Jacobian of exponential map w.r.t. $\uv$ then follows
	\begin{equation}\label{eq:dexp}
	 \pad{\Exp_\dso(\uv)}{\uv}\lsft{-3mu}=\lsft{-3mu}\Big[\Big(\pad{(\cos\nm{\uv})}{\uv}\Big)^\top,\Big(\pad{(\uv\sinc\nm{\uv})}{\uv}\Big)^\top\Big]^\top\lsft{-15mu}\in\R^{4\times{3}}\,,
	\end{equation}
	 with the two items expressed as 
	 \begin{equation*}
	 	\begin{aligned}
 		&\pad{(\cos\nm{\uv})}{\uv}=-\uv^\top\sinc\nm{\uv}\eqand\\
		&\pad{(\uv\sinc\nm{\uv})}{\uv}=\frac{\cos\nm\uv-\sinc\nm\uv}{\nm{\uv}^2}\uv\,\uv^\top+\eye{3}\sinc\nm\uv\,.
	 	\end{aligned}
	 \end{equation*}

	\subsection{Jacobian of quaternion logarithm map}
	Given any unit quaternion $\ur=[\,q_0,\ur_\ttv^\top]^\top\in\Sbb^3$ with $q_0$ and $\ur_\ttv$ denoting the scalar and vector components, the rotation angle and axis can be retrieved by definition as $\theta=2\arctan(\nm{\ur_\ttv}/q_0)$ and $\uu=\ur_\ttv/\nm{\ur_\ttv}$, respectively. It can be mapped to the tangent space at $\dso$ via logarithm map
	\begin{equation*}
		\Log_\dso(\ur)=\theta\uu/2=\arctan(\nm{\ur_\ttv}/q_0)\ur_\ttv/\nm{\ur_\ttv}\in\R^3\,.
	\end{equation*}
	The Jacobian of logarithm map follows
	\begin{equation}\label{eq:dlog}
	\pad{\Log_\dso(\ur)}{\ur}=\bigg[\pad{\Log_\dso(\ur)}{q_0},\pad{\Log_\dso(\ur)}{\ur_\ttv}\bigg]\in\R^{3\times{4}}\,,
	\end{equation}
 	\begin{equation*}
		\begin{aligned}
		&\text{where}\quad \pad{\Log_\dso(\ur)}{q_0}=-\ur_\ttv\eqand\\
		&\pad{\Log_\dso(\ur)}{\ur_\ttv}=\frac{1}{\nm{\ur_\ttv}^2}\Big(q_0\ur_\ttv\ur_\ttv^\top+\frac{\nm{\ur_\ttv}^2\eye{3}-\ur_\ttv\ur_\ttv^\top}{\nm{\ur_\ttv}}\arctan\frac{\nm{\ur_\ttv}}{q_0}\Big)\,.
		\end{aligned}
	\end{equation*}
	
	\section*{Acknowledgment}
	We would like to thank Norbert Fränzel from Fraunhofer IOSB-AST and Christopher Funk for help with instrumenting the sensor suite to record \ttt{ISAS-Walk} data set.

	\bibliographystyle{IEEEtran.bst}
	\bibliography{bibliography.bib}
	
\end{document}